\PassOptionsToPackage{table,dvipsnames}{xcolor}
\documentclass[]{selfevolagent}

\usepackage{microtype}
\usepackage{amsfonts}
\usepackage{amsmath}
\usepackage{xcolor}
\usepackage{graphicx}
\usepackage{booktabs}
\usepackage{tabularx}
\usepackage{caption}
\usepackage{url}
\usepackage[most]{tcolorbox}
\usepackage{fontawesome5}
\usepackage[T1]{fontenc}

\title{AREX-2: Advancing Self-Improving Agents\\
through Long-Horizon Reflective Tasks}

\author{AREX Team}

\affiliation{Beijing Academy of Artificial Intelligence (BAAI)}

\abstract{
We present \textbf{AREX-2}, an effort to advance the self-improving
capability of LLM agents, which we define as the ability to iteratively
refine a solution at test time. This ability rests on two complementary
capabilities: \emph{reflection}, which produces a solution better than
the current one, and \emph{long-horizon execution}, which keeps the
iteration effective over many rounds. We hypothesize that both
capabilities are domain-agnostic, and can therefore be learned in
scenarios that are well suited for supervision. Accordingly, we
synthesize long-horizon improvement trajectories from machine learning
and algorithmic programming tasks, two domains that offer verifiable
feedback and reward sustained iteration. Trained on this data, our agent,
built on Qwen3.8-27B, achieves strong results on MLE-bench~Lite~(81.8) and
Frontier-CS~(70.7), transfers to deep research with 84.0 on BrowseComp,
52.6 on HLE, 92.2 on GAIA, and 93.8 on DeepSearchQA, and keeps improving
as its budget of rounds grows. These results show that long-horizon reflective data is an
effective route toward self-improving agents.
}

\metadata[{\raisebox{-0.2ex}{\includegraphics[height=1em]{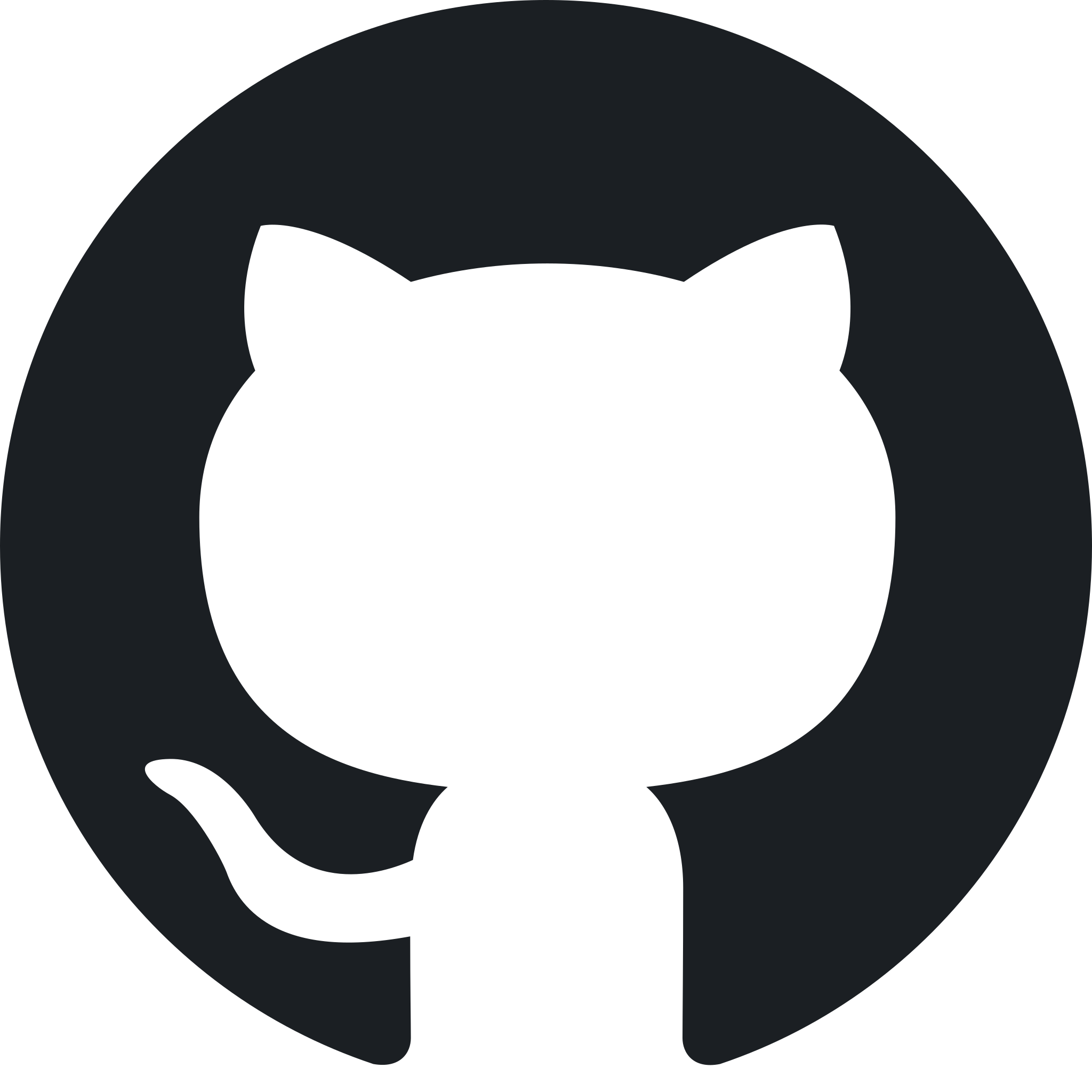}}\ Homepage}]{\url{https://github.com/VectorSpaceLab/AREX-2}}
\metadata[{\raisebox{-0.2ex}{\includegraphics[height=1em]{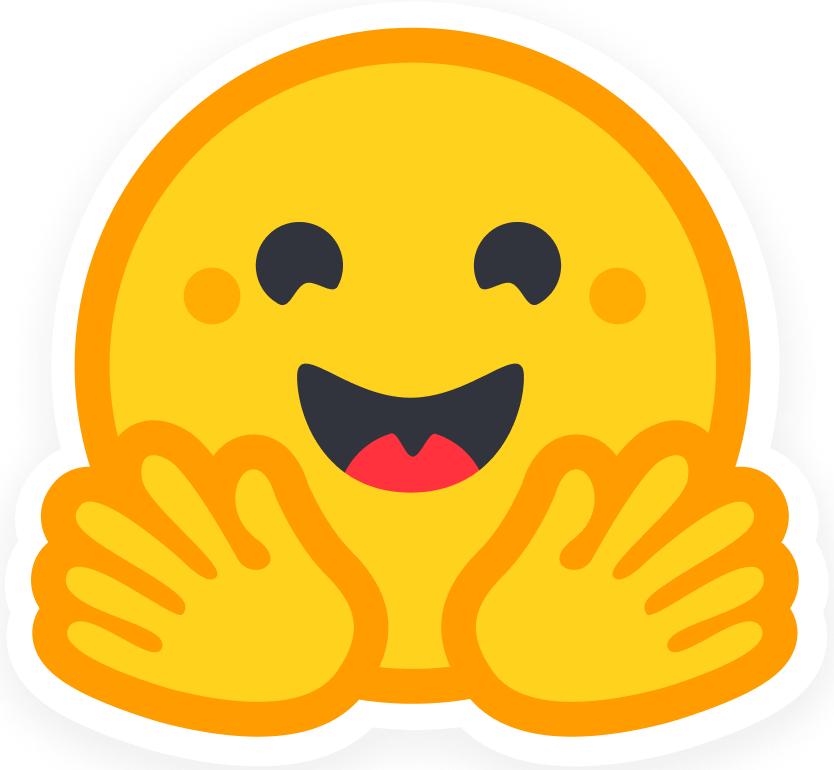}}\ Models}]{\url{https://huggingface.co/collections/BAAI/arex-2}}
\metadata[\faEnvelope\ Correspondence]{chienqhj@gmail.com, zhengliu1026@gmail.com}

\AtBeginDocument{\setlength{\parfillskip}{0pt plus 0.75\textwidth}\setlength{\emergencystretch}{1.5em}}

\begin{document}

\maketitle

\begin{figure}[!ht]
    \centering
    \includegraphics[width=\linewidth]{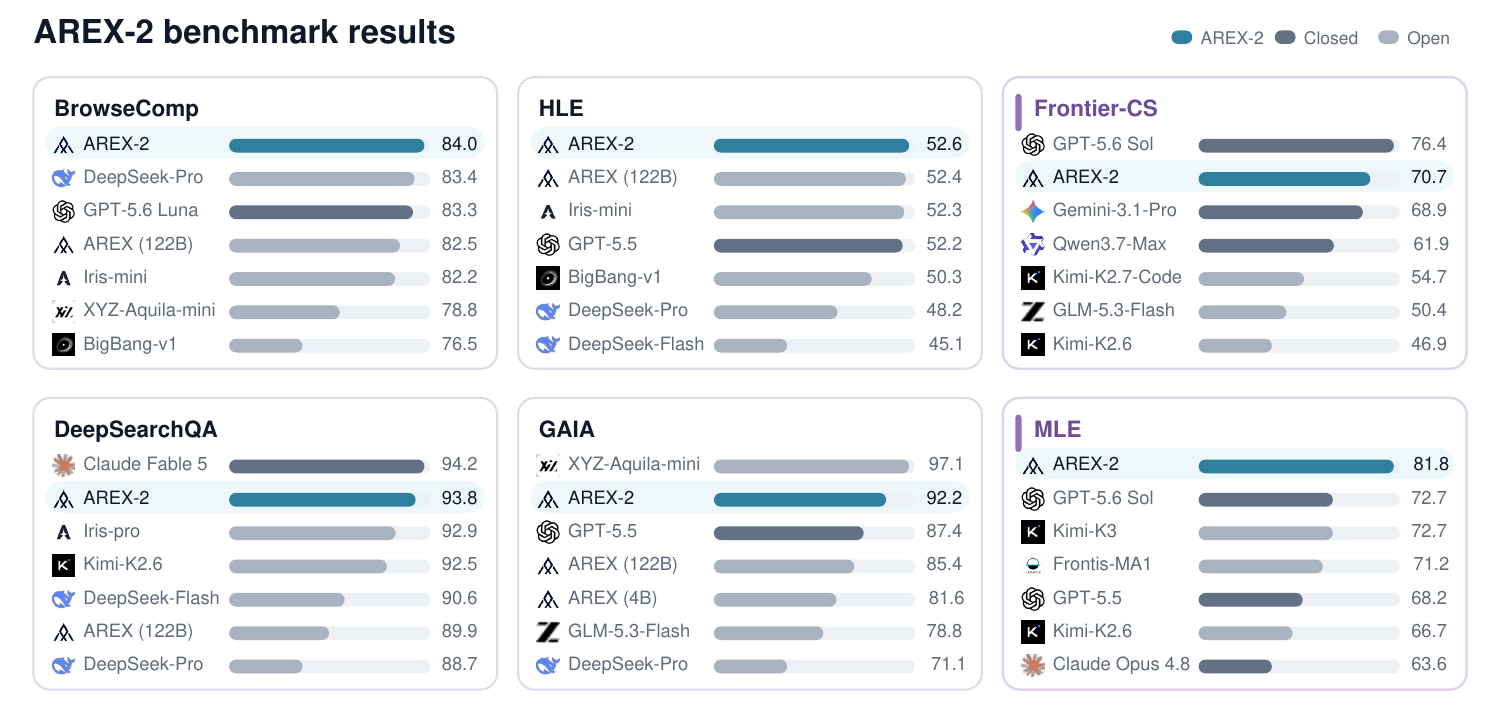}
    \caption{Results of AREX-2 on six benchmarks, compared with selected closed and open models.}
    \label{fig:benchmark}
\end{figure}

\vspace{-8pt}
\begin{figure}[!ht]
    \centering
    \includegraphics[width=0.95\linewidth]{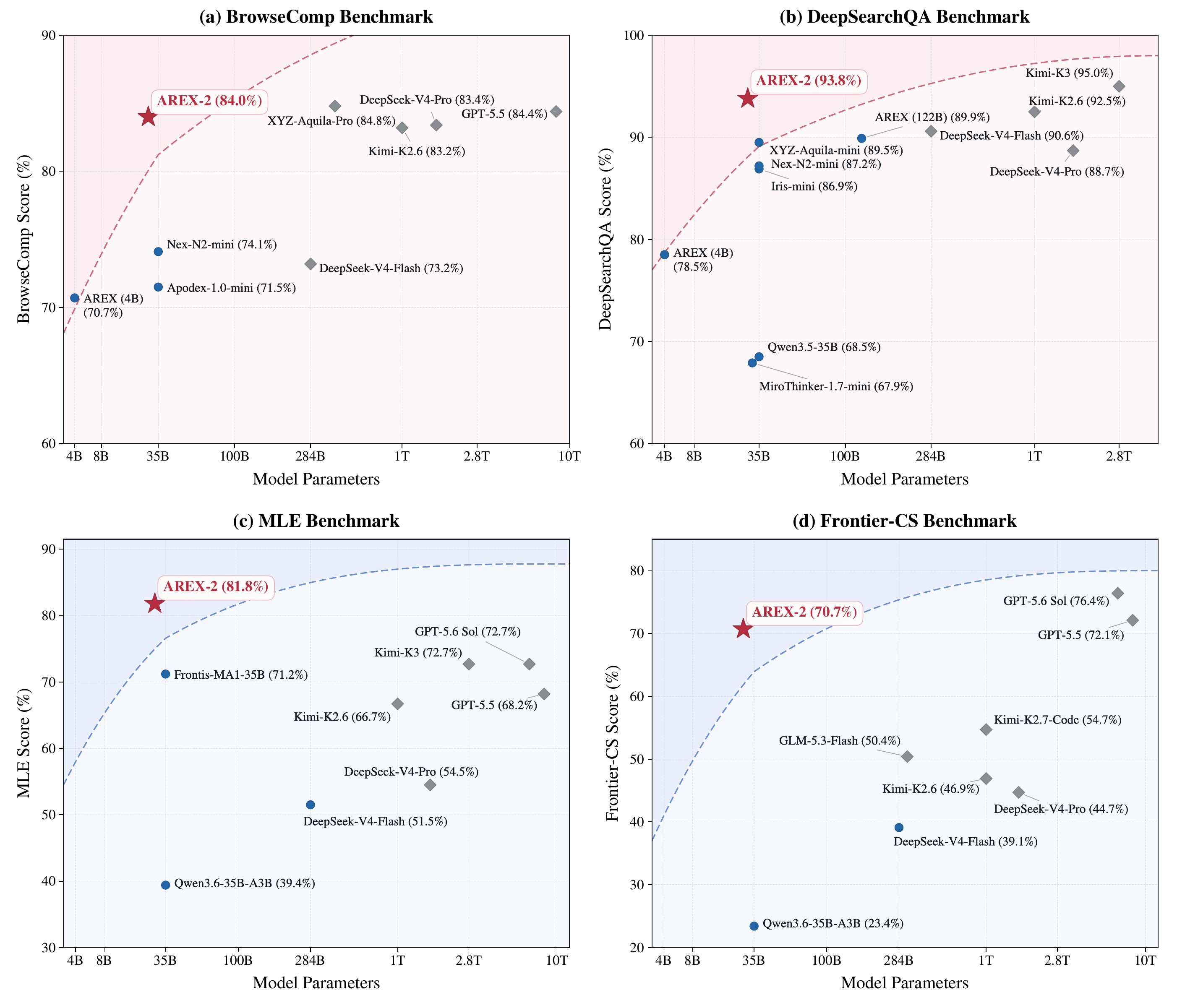}
    \vspace{-8pt}
    \caption{Model size versus performance on four benchmarks. AREX-2 (27B) matches or exceeds much larger models.}
    \label{fig:cost}
    \vspace{-8pt}
\end{figure}

\section{Introduction}
\label{sec:intro}

Hard problems are not solved in one attempt. A training pipeline, a program under a scoring function, or a research question yields to a loop of trying, measuring, and revising, and the systems that do such work well are built around that loop~\citep{aide2025,alphaevolve2025,aiscientist2024,frontis2026}. We call one pass through the loop a \emph{round}. In most of these systems, however, the loop is implemented largely in the scaffold rather than in the model. The harness decides when to retry and what to keep, while the model is asked mainly to produce one attempt at a time. This paper asks how to move the loop inside the model, and defines \emph{self-improvement} accordingly: the ability of an agent, given more rounds on a task, to turn them into a better solution by its own judgment of what to change. Self-improvement in this sense is test-time scaling within a single task: the more rounds the agent spends, the better its solution should become.

Whether more rounds help depends on two capabilities. \emph{Reflection} is producing, from the current solution and what the feedback says about it, a better one. \emph{Long-horizon execution} is sustaining this process across rounds, retaining useful findings and recovering from setbacks. Let $s_t$ be the best score the agent has reached after $t$ rounds and $r_t = \mathbb{E}[s_t-s_{t-1}]$ the expected gain of round $t$. From a fixed initial score $s_0$ and a budget of $T$ rounds, the expected total improvement is
\begin{equation}
  \mathbb{E}[s_T] - s_0 = \sum_{t=1}^{T} r_t .
  \label{eq:self-improvement}
\end{equation}
Let $T^{\ast} \le T$ be the number of rounds in which $r_t$ remains non-negligible, and $\bar r$ the mean gain over those rounds, so that the total improvement is approximately $\bar r\,T^{\ast}$. Reflection determines $\bar r$, how much a productive round is worth; long-horizon execution determines $T^{\ast}$, how many rounds stay productive before the agent stalls. The two capabilities are therefore complementary: when $\bar r$ is small, further rounds add little; when $T^{\ast}$ is short, even sharp reflection has limited cumulative effect, and a larger budget helps only up to $T^{\ast}$. A self-improving agent needs both, and we refer to the two together as \emph{long-horizon reflection}.

To move the loop inside the model, both capabilities have to be learned in training, and a model learns mainly what its training data demonstrates. Current training data for agents demonstrates little of either. It is typically built one attempt at a time: a task is posed, the model produces a solution, a verifier marks it pass or fail, and the passing solutions are kept for training~\citep{swegym2025,swesmith2025,r2egym2025}. This follows the general practice of fine-tuning a model on its own successful outputs~\citep{star2022,rest2023,rft2023}. Data built this way shows a model what a correct solution looks like, but not how a solution is improved. The intermediate attempts, the feedback they received, and the revisions that followed are all discarded, and because solutions that pass immediately are the easiest to collect, long processes of improvement are the first to be lost. As a result, trajectories in which an agent works on one problem for hours, measuring and revising until it ends far above where it began, are almost absent from training data. Such trajectories have to be constructed, which raises two questions: in which domains to construct them, and how.

For the first question, we start from a hypothesis: \textbf{long-horizon reflection is a meta-skill that is not tied to any one domain.} Judging where a solution falls short, deciding what to change, and continuing after a failed round are the same acts whether the solution is a training pipeline, a program, or a research report. If the hypothesis holds, the training domain need not resemble the target domain. It should instead be chosen for how well it allows the meta-skill to be supervised, and what is learned there should transfer to other domains. By this criterion we choose two domains, machine learning engineering and algorithmic programming, for three reasons. First, their feedback is unambiguous: a validation score or a judge's verdict shows directly whether a revision helped. Second, they leave room for sustained improvement: a solution can keep getting better over many rounds, from a baseline to a tuned ensemble or from a brute-force program to a near-optimal one, so long trajectories are rewarded~\citep{mlebench2025,frontiercs2025}. Third, their source material is abundant: GitHub and online judges offer a large supply of problems, code, and tests. Together, these properties make the two domains well suited to supervising self-improvement.

For the second question, we construct the data in three steps. First, a teacher model turns source material from GitHub and online judges into executable environments, each consisting of a task and a scoring function. Second, an agent works on each task over many rounds, revising its solution according to the feedback it receives. Third, we select trajectories as a whole: a trajectory is kept if its final score is high and its process satisfies the task's requirements, and no individual round is required to succeed. This choice is deliberate. A trajectory selected by its final outcome still contains failed runs, regressions, and abandoned approaches, which filtering at the level of individual steps would remove. Keeping them teaches the model to continue iterating after a setback. Removing them would leave the model with only examples in which every step succeeded, and with nothing to learn from when a round fails.

We train \textbf{AREX-2} from Qwen3.8-27B~\citep{qwen38} on this data, together with the deep-research data of the \textsc{AREX} recipe~\citep{arex2026}. In the two training domains, it reaches 81.8 on MLE-bench Lite, the highest score among the systems we compare, and 70.7 on Frontier-CS, the highest among open-weight models (\Cref{fig:benchmark}). In deep research, for which we added no new training data, it still improves over the previous \textsc{AREX} models, reaching 84.0 on BrowseComp, 52.6 on HLE, 92.2 on GAIA, and 93.8 on DeepSearchQA: long-horizon reflection carries over beyond the domains it was learned in. With only 27B parameters, AREX-2 compares favorably with substantially larger models (\Cref{fig:cost}), and it keeps improving as its budget of rounds grows.

\section{Method}
\label{sec:method}

\begin{figure}[!ht]
    \centering
    \includegraphics[width=\linewidth]{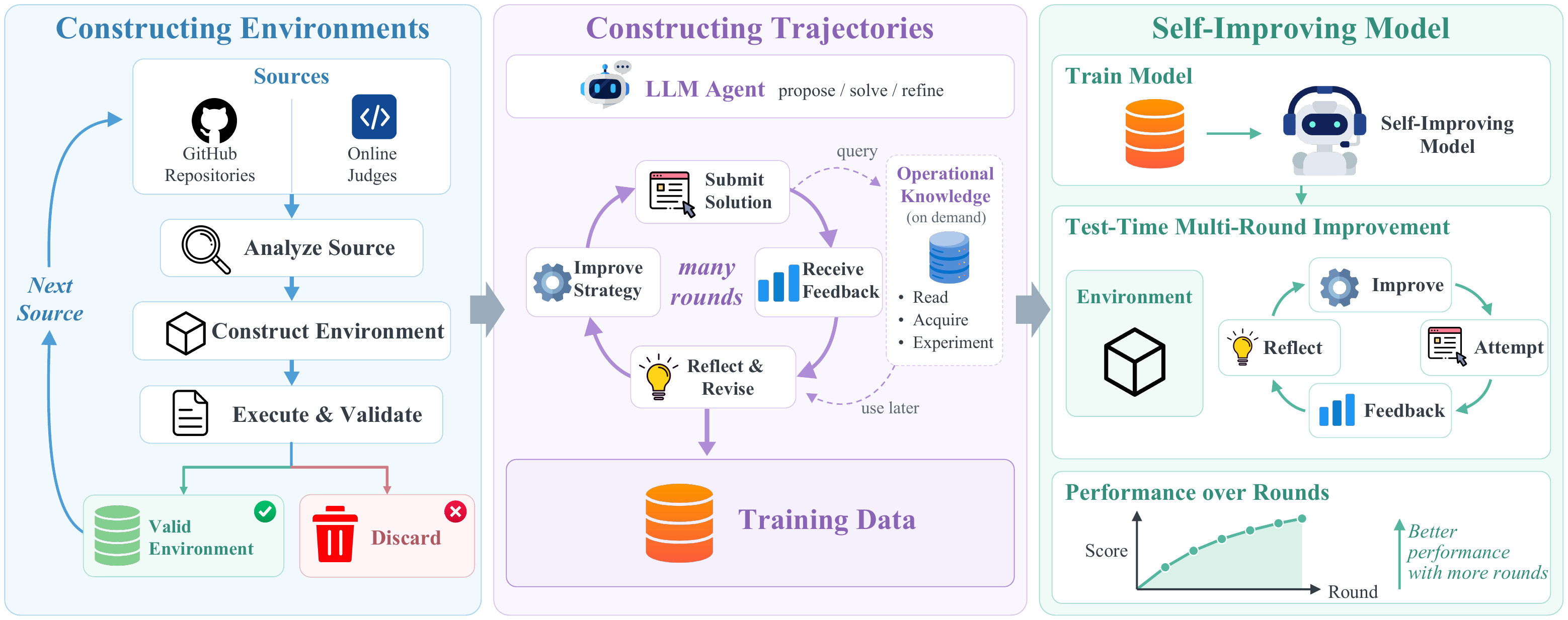}
    \caption{Overview of AREX-2. \textbf{Left:} environments are constructed from GitHub repositories and online judges, and are kept only if they pass execution and validation. \textbf{Middle:} an agent improves its solution over many rounds, acquiring operational knowledge on demand, and its trajectories become training data. \textbf{Right:} the model trained on this data improves its solution over multiple rounds at test time.}
    \label{fig:overview}
\end{figure}

\subsection{Self-Improvement as Test-Time Scaling}
\label{sec:formalization}

An environment is a pair $e = (\sigma, S)$, where $\sigma$ is the task given to the agent and $S$ is a scoring function that maps a candidate solution $y$ to a scalar $S(y)$. Examples of $S$ are the validation metric of a trained model and the fraction of hidden tests a program passes; we orient $S$ so that higher is better. The agent works in $e$ over rounds. In round $t$ it submits a solution $y_t$ and receives feedback $f_t$, which consists of the score and whatever else the environment reports, such as logs, errors, and timings. A round includes everything the agent does before the submission: reading, editing, running, and searching. A trajectory of $n$ rounds is
\begin{equation}
  \tau \;=\; \big(y_0, f_0,\; y_1, f_1,\; \ldots,\; y_n, f_n\big),
  \qquad s_t = \max_{k \le t} S(y_k),
\end{equation}
where $s_t$ is the best score reached after $t$ rounds. Under a policy $\pi_\theta$, the expected gain of round $t$ is
\begin{equation}
  r_t \;=\; \mathbb{E}_{\pi_\theta}[s_t-s_{t-1}].
\end{equation}
For a fixed initial score $s_0$ and a budget of $T$ rounds, the expected total improvement is therefore
\begin{equation}
  \mathbb{E}_{\pi_\theta}[s_T]-s_0 = \sum_{t=1}^{T} r_t,
\end{equation}
as in \Cref{eq:self-improvement}. The budget $T$ is the resource scaled at test time. It is given to the agent, not learned by it; what the policy determines is how much of the budget it can use productively. For a small threshold $\epsilon$, let
\begin{equation}
  T^{\ast} \;=\; \big|\{\,t \le T : r_t > \epsilon\,\}\big|,
  \qquad
  \bar r \;=\; \frac{1}{T^{\ast}} \sum_{t \le T:\; r_t > \epsilon} r_t
\end{equation}
be the number of rounds that remain productive and the mean gain over those rounds. The total improvement is then $\bar r\,T^{\ast}$, up to an error of at most $\epsilon T$. Reflection determines $\bar r$ and long-horizon execution determines $T^{\ast}$, so a larger budget helps only up to $T^{\ast}$. Because $s_t$ is the best score so far, a single submission may score below it without lowering $s_t$, and $r_t$ decreases as the score approaches its ceiling.

An agent is self-improving in $e$ to the degree that more rounds lead to better solutions in expectation. Our hypothesis is that long-horizon reflection is a meta-skill: the behaviors that produce such improvement, which are using feedback, acquiring operational knowledge, and recovering from setbacks, transfer across families of environments even when their tasks and scoring functions differ.

We train $\pi_\theta$ by imitation, on trajectories that exhibit these behaviors. The rest of this section describes how the trajectories are obtained: how environments are built (\Cref{sec:environments}), how an agent is run in them to produce trajectories of sustained improvement (\Cref{sec:trajectories}), and how the trajectories are selected and turned into supervision (\Cref{sec:training}).

\subsection{Constructing Environments}
\label{sec:environments}

We build environments from two kinds of source: GitHub repositories, which supply machine learning tasks, and online judges, which supply algorithmic programming tasks. A source is not yet an environment. A repository has code, data, and a way of measuring a model, but it does not state a task. A judge problem has a statement and tests, but it has no sandbox in which an agent can iterate. A teacher model therefore turns each source into an environment $(\sigma, S)$.

For a repository, the teacher reads the code and identifies the metric the repository measures. It then writes a task that asks the agent to improve this metric, a scoring script that computes the metric on a held-out split the agent cannot access, and a sandbox with the repository installed and its data in place. For a judge problem, the task is the problem statement, the sandbox contains a compiler and the sample cases, and $S$ is the score assigned by the hidden tests. Wherever the judge allows it, this score is graded, not binary: for example, the fraction of tests passed, or the quality of a heuristic solution relative to the best known one. A graded score makes each round informative, whereas a binary verdict says only that the agent has not succeeded.

An environment is admitted if it satisfies two conditions, both checked by execution. First, the reference solution that comes with the source, which is the repository's own model or the judge's accepted submission, must run under $S$ and obtain a score. This shows that the environment works. Second, a simple baseline, which is the repository as found or a straightforward first program, must score well below the reference. This shows that the environment leaves room for improvement. Environments that fail the second condition are dropped, because they cannot produce trajectories of sustained improvement.

\subsection{Constructing Trajectories}
\label{sec:trajectories}

A trajectory is produced by running a teacher agent in an environment, under three conditions designed to elicit long-horizon reflection.

\paragraph{More rounds.} The agent is given a budget of rounds and wall-clock time far larger than a single attempt needs, on the order of hours and hundreds of tool calls per task, and it is told what the budget is. Knowing the budget changes how the agent works. It can establish a baseline first, use early rounds to measure, and use later rounds for the revisions that the measurements suggest, instead of submitting its best guess at once. The trajectory thus demonstrates how to turn a larger round budget into cumulative improvement, and the round budget is the same resource that we scale at test time.

\paragraph{Operational knowledge.} To improve a solution, the agent has to know how the environment works: which API the repository exposes, what its data loader expects, which optimizations the constraints of a problem permit. We call this \emph{operational knowledge}. It is not given in the task statement, so the agent acquires it while working, by reading the repository's source and documentation, searching for the papers and libraries that the code depends on, and running small experiments in the environment. The agent then uses what it has learned in the rounds that follow. All of these actions remain in the trajectory, so a model trained on the trajectory learns to investigate how an environment works as part of improving a solution. Part of this knowledge can be provided as \emph{skills}, which are compact documents placed in the agent's context. General skills are prepared in advance from machine learning repositories and common practices. Task-specific skills are built by searching for information related to a task, and material about the task itself is excluded. We provide skills to the agent that produces the trajectories, because they make it capable enough to produce trajectories worth imitating. The trained model learns from these trajectories how to use such knowledge, and skills remain available to it at test time.

\paragraph{Feedback in the loop.} Each round ends with a submission, and the agent sees the score and the environment's report before it decides what to do next. This is the point at which reflection takes place: the agent compares the result with what it expected, keeps or reverts its change, and forms its next hypothesis. Rounds that lower the score are not removed from the trajectory. Each is followed by the agent's response to it, and this response is what we want the model to learn.

For each environment, the result is a set of trajectories in which a capable agent works on one task over many rounds, with its reasoning, searches, failures, and recoveries all recorded.

\subsection{Training}
\label{sec:training}

\paragraph{Trajectory selection.} A trajectory is kept or discarded as a whole, according to two conditions. First, its final score must be high: the best submission must reach a threshold set relative to the reference, so that the trajectory demonstrates a real improvement. Second, its process must satisfy the task's requirements: the transcript follows the user-assistant-tool format, every tool call has an observation, the run terminates in a well-formed state, and the agent has not violated the rules of the task, for example by reconstructing answers from the held-out split or exposing credentials. We place no condition on individual rounds. A kept trajectory may therefore contain failed runs, regressions, and abandoned approaches. We keep them deliberately, because they show the model how to continue after a setback.

\paragraph{Supervision within a trajectory.} The whole trajectory is kept as context, and the loss is applied to the decisions that move the solution forward. We group the agent's outputs into steps, where a step is one decision together with the actions issued before the next observation. Failed attempts and regressions stay in the context, so that the model sees the state from which it has to recover. The loss is applied to the decisions that follow them and make progress: diagnosing the failure, repairing it, changing strategy, running the next experiment, and submitting an improved solution.

Steps that make no progress receive no loss. These include repeated polling, calls that return no new observation, and near-duplicate turns. System messages, environment reports, and retrieved documents also receive no loss, since they condition the model but are not outputs to imitate. The model thus learns how to respond to a failure and turn it into a later improvement, while the failure itself stays in the context.

\paragraph{Data and model.} We combine the selected trajectories with the deep-research data of the \textsc{AREX} recipe~\citep{arex2026}, which we leave unchanged, and fine-tune Qwen3.8-27B~\citep{qwen38} on the mixture to obtain AREX-2. The new trajectories are the only difference from the previous recipe. The cross-domain results in \Cref{sec:experiments} can therefore be read as transfer: any gain of AREX-2 on deep research over models trained with the previous recipe does not come from new deep-research data.

\section{Experiments}
\label{sec:experiments}

Our experiments ask whether long-horizon improvement trajectories
produce strong performance in their source domains, whether the learned
capabilities extend to deep research, and whether the trained agent
turns additional test-time computation into continued progress. We
first report overall benchmark results, then examine how the agent
improves during execution and how the data-generation recipe teaches
that behavior.

\subsection{Experimental Setup}

\paragraph{Benchmarks.} We evaluate AREX-2 on six benchmarks covering four capabilities: algorithmic programming, machine learning engineering, deep research, and general agentic reasoning. \textbf{Frontier-CS}~\citep{frontiercs2025} evaluates algorithmic programming on open-ended problems with graded scores, where an agent improves its solution through repeated implementation, testing, and submission. \textbf{MLE-bench Lite}~\citep{mlebench2025} (MLE-Lite for short) measures machine learning engineering capabilities, where agents must complete end-to-end ML workflows involving data analysis, experimentation, implementation, and model evaluation. For deep research and agentic reasoning, we include \textbf{BrowseComp}~\citep{browsecomp2025}, \textbf{HLE}~\citep{hle2025}, \textbf{GAIA}~\citep{gaia2023}, and \textbf{DeepSearchQA}~\citep{deepsearchqa2026}. These benchmarks cover deep research, tool-augmented reasoning, information gathering, and multi-step task completion. Following standard evaluation protocols, we report F1 for \textbf{DeepSearchQA}, Any Medal for \textbf{MLE-Lite}, and accuracy for the remaining benchmarks. Any Medal is the percentage of competitions in which the agent earns a medal, averaged over three seeds.

\paragraph{Evaluation protocols.} For non-coding benchmarks, we follow the evaluation protocol of \textbf{\textsc{AREX}}~\citep{arex2026}, setting the maximum number of inner turns to 300 and the overall maximum number of turns to 1500. For coding-related evaluations, we adopt the official agent-based evaluation settings of each benchmark. Specifically, \textbf{Frontier-CS} is evaluated under the Agent Track with a maximum execution budget of 5 hours per task, while \textbf{MLE-Lite} follows the OpenMLE evaluation protocol~\citep{frontis2026} with a maximum budget of 12 hours per task. On MLE-Lite, AREX-2 is evaluated with skills in its context; \Cref{sec:case-mle} reports how much the skills and the training each contribute. All evaluations are conducted using the corresponding benchmark environments, allowing agents to iteratively reason, act, and refine their solutions within the budget.

\begin{table*}[t]
  \centering
  \begin{minipage}{0.68\textwidth}
  \centering

  {\small
  \renewcommand{\arraystretch}{1.12}
  \setlength{\tabcolsep}{7.5pt}

  \begin{tabularx}{\linewidth}{
    @{} l c
    >{\raggedleft\arraybackslash}X
    >{\raggedleft\arraybackslash}X
    @{}
  }
  \toprule
  \textbf{Model} & \textbf{Size} & \textbf{Frontier-CS} & \textbf{MLE-Lite} \\
  \midrule

  \multicolumn{4}{@{}l}{\textit{Closed-Weight Models}} \\
  GPT-5.6 Sol       & -- & 76.4 & 72.7 \\
  Claude Opus 4.8   & -- & 74.5 & 63.6 \\
  GPT-5.5           & -- & 72.1 & 68.2 \\
  Gemini-3.1-Pro    & -- & 68.9 & -- \\
  Qwen3.7-Max       & -- & 61.9 & -- \\

  \midrule
  \multicolumn{4}{@{}l}{\textit{Open-Weight Models}} \\
  Kimi-K3                         & 2.8T & --          & 72.7 \\
  Naive-N0.5-Flash                & 309B & --          & 73.7 \\
  DeepSeek-V4-Pro                 & 1.6T & 44.7\rlap{$^{*}$} & 54.5 \\
  DeepSeek-V4-Flash               & 284B & 39.1\rlap{$^{*}$} & 51.5 \\
  Kimi-K2.7-Code                  & 1T   & 54.7\rlap{$^{*}$} & -- \\
  GLM-5.3-Flash                   & 320B & 50.4\rlap{$^{*}$} & -- \\
  Kimi-K2.6                       & 1T   & 46.9        & 66.7 \\
  Frontis-MA1-35B                 & 35B  & --          & 71.2 \\
  BigBang-V1                      & 35B  & --          & 59.1 \\
  Qwen3.6-35B-A3B                 & 35B  & 23.4\rlap{$^{*}$}          & 39.4 \\

  \midrule
  \rowcolor{blue!5}
  \textbf{AREX-2} & \textbf{27B} & \textbf{70.7} & \textbf{81.8} \\
  \bottomrule
  \end{tabularx}
  }

  \end{minipage}

  \caption{Comparison on coding and machine learning engineering benchmarks. Model size refers to the total number of parameters. Frontier-CS results are reported on the 188-task Agent Track. $^{*}$ denotes results reproduced by us. MLE-Lite results are reproduced and reported by OpenMLE. MLE-Lite reports Any Medal (mean over three seeds) on MLE-bench Lite. AREX-2 is evaluated on MLE-Lite with skills in its context (\Cref{sec:case-mle}).}
  \label{tab:coding-mle-comparison}
\end{table*}

\begin{table*}[t]
    \centering
    \begin{minipage}{0.88\textwidth}
    \centering

    {\small
    \renewcommand{\arraystretch}{1.12}
    \setlength{\tabcolsep}{4.5pt}

    \begin{tabularx}{\linewidth}{
        @{} l c
        >{\raggedleft\arraybackslash}X
        >{\raggedleft\arraybackslash}X
        >{\raggedleft\arraybackslash}X
        >{\raggedleft\arraybackslash}X
        @{}
    }
    \toprule
    \textbf{Model} & \textbf{Size} & \textbf{BrowseComp} & \textbf{HLE} & \textbf{GAIA} & \textbf{DeepSearchQA} \\
    \midrule

    \multicolumn{6}{@{}l}{\textit{Frontier Models}} \\
    GPT-5.6 Sol       & --   & 90.4 & 58.0\rlap{\textsuperscript{*}} & --   & --   \\
    GPT-5.6 Terra     & --   & 87.5 & -- & --   & --   \\
    GPT-5.6 Luna      & --   & 83.3 & -- & --   & --   \\
    Kimi-K3           & 2.8T & 91.2 & 56.0\rlap{\textsuperscript{*}} & --   & 95.0 \\
    Claude Fable 5    & --   & 88.0 & 64.5\rlap{\textsuperscript{*}} & --   & 94.2 \\
    Claude Opus 4.8   & --   & 84.3 & 57.9\rlap{\textsuperscript{*}} & --   & 93.1 \\
    GPT-5.5           & --   & 84.4 & 52.2\rlap{\textsuperscript{*}} & 87.4 & --   \\
    Gemini-3.1-Pro    & --   & 85.9 & 51.4\rlap{\textsuperscript{*}} & 80.6 & 93.3 \\

    \midrule
    \multicolumn{6}{@{}l}{\textit{Large Models ($>$40B)}} \\
    GLM-5             & 744B & 75.9 & 50.4 & 70.0 & --   \\
    Kimi-K2.6         & 1T   & 83.2 & 54.0\rlap{\textsuperscript{*}} & 80.6 & 92.5 \\
    GLM-5.3-Flash     & 320B &  --  & 55.3\rlap{\textsuperscript{*}} & 78.8 & -- \\
    DeepSeek-V4-Flash & 284B & 73.2 & 45.1 & 57.5   & 90.6 \\
    DeepSeek-V4-Pro   & 1.6T & 83.4 & 48.2 & 71.1   & 88.7 \\
    MiroThinker-1.7   & 235B & 74.0 & 42.9 & 82.7 & 72.1 \\
    XYZ-Aquila-pro    & 397B & 84.8 & 53.3 & -- & 92.5 \\
    Iris-pro          & 397B & 88.6 & 56.4 & --   & 92.9 \\
    \textsc{AREX} (122B) & 122B & 82.5 & 52.4 & 85.4 & 89.9 \\

    \midrule
    \multicolumn{6}{@{}l}{\textit{Small Models ($\leq$40B)}} \\
    Tongyi-DeepResearch-30B & 30B & 43.4 & 32.9 & 70.9 & --   \\
    Qwen3.5-35B             & 35B & 61.0 & 47.4 & 80.0 & 68.5 \\
    XYZ-Aquila-mini         & 35B & 78.8 & 51.1 & 97.1 & 89.5 \\
    BigBang-V1              & 35B & 76.5 & 50.3 & --   & --   \\
    Quest-35B               & 35B & 64.6 & 37.2 & 80.8 & --   \\
    Apodex-1.0-mini         & 35B & 71.5 & 46.8 & --   & 82.2 \\
    Agents-A1               & 35B & 75.5 & 47.6 & 96.0 & --   \\
    MiroThinker-1.7-mini    & 30B & 67.9 & 36.4 & 80.3 & 67.9 \\
    Iris-mini               & 35B & 82.2 & 52.3 & --   & 86.9 \\
    \textsc{AREX} (4B)     & 4B  & 70.7 & 40.6 & 81.6 & 78.5 \\

    \midrule
    \rowcolor{blue!5}
    \textbf{AREX-2} & \textbf{27B} & \textbf{84.0} & \textbf{52.6} & \textbf{92.2} & \textbf{93.8} \\
    \bottomrule
    \end{tabularx}
    }

    \end{minipage}

    \caption{Comparison on general agentic reasoning and deep research benchmarks. $^\ast$ denotes results reported on the full HLE set; unmarked results use the text-only subset.}
    \label{tab:agentic-benchmark}
\end{table*}

\subsection{Overall Results}
\label{sec:experimental-results}

\paragraph{Learning to solve verifiable tasks.}
\Cref{tab:coding-mle-comparison} evaluates performance in the domains
used to construct the new training trajectories. AREX-2 reaches an Any Medal rate of 81.8
on MLE-bench Lite, the highest score in the table: 8.1 points above the
strongest baseline, Naive-N0.5-Flash, and 9.1 points above the strongest
closed-weight baseline, GPT-5.6 Sol. On Frontier-CS, it achieves 70.7,
exceeding the strongest reported open-weight baseline by 16.0 points
and placing within 5.7 points of the strongest closed-weight system.
At 27B parameters, AREX-2 therefore combines leading performance on
machine learning engineering with competitive performance on
algorithmic programming among the systems compared here.

\paragraph{Extending capabilities without new research data.}
Table~\ref{tab:agentic-benchmark} evaluates cross-domain transfer: no newly constructed trajectory is a search task, while the deep-research training data remains unchanged from the previous \textsc{AREX} recipe. AREX-2 obtains 84.0 on BrowseComp, 52.6 on
text-only HLE, 92.2 on GAIA, and 93.8 on DeepSearchQA. It leads the
reported models with at most 40B parameters on BrowseComp, text-only
HLE, and DeepSearchQA; on GAIA, it ranks behind XYZ-Aquila-mini and
Agents-A1 but above the remaining small models with reported results.
Its performance is also competitive with substantially larger
systems: it exceeds DeepSeek-V4-Pro and Kimi-K2.6 on BrowseComp and
ranks third on DeepSearchQA among the models in the table.
Most importantly, AREX-2 surpasses both models trained with the
previous recipe, \textsc{AREX} (4B) and \textsc{AREX} (122B),
on all four benchmarks.

\paragraph{Overall.} AREX-2 combines strong performance in machine learning engineering and algorithmic programming with substantial gains in deep research over the previous \textsc{AREX} models, although it uses the same deep-research training data. This pattern is consistent with the hypothesis in \Cref{sec:intro} that long-horizon reflection is a meta-skill: what the model learns from the two training domains carries over beyond them. A gap nevertheless remains on the strongest BrowseComp and HLE results.

\subsection{Detailed Analysis}
\label{sec:cases}

The results in \Cref{sec:experimental-results} show what AREX-2 achieves, but not how. This section examines two questions that follow from the formalization in \Cref{sec:formalization}. The first is whether AREX-2 turns a larger budget into a better solution, that is, whether its effective horizon $T^{\ast}$ is long. We test this in two settings: one in which the agent receives a score after each submission, and one in which it receives no feedback on correctness. The second question is how much each part of our approach contributes: operational knowledge, training, and a larger budget. We test this with a stage-wise ablation.

\subsubsection{Round Scaling with Feedback}
\label{sec:case-fcs}

\begin{figure*}[t]
  \centering

  \begin{minipage}[t]{0.485\textwidth}
    \centering
    \includegraphics[width=\linewidth]{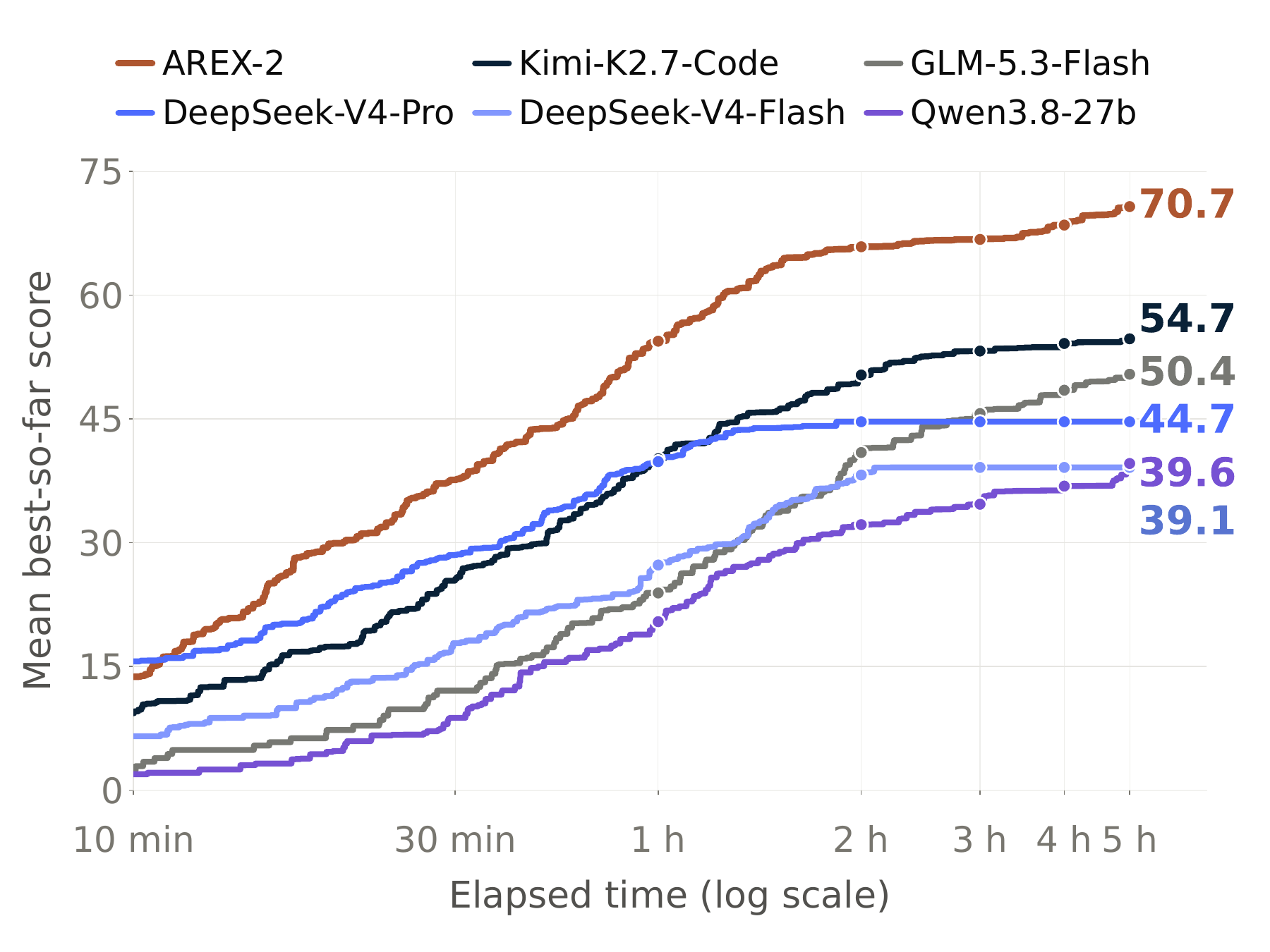}
    \captionof{figure}{
      Test-time scaling for agentic coding tasks.
      Mean best-so-far score across 188 Frontier-CS Agent Track
      problems. Elapsed time starts at each task's agent session.
    }
    \label{fig:fcs-main}
  \end{minipage}
  \hfill
  \begin{minipage}[t]{0.488\textwidth}
    \centering
    \includegraphics[width=\linewidth]{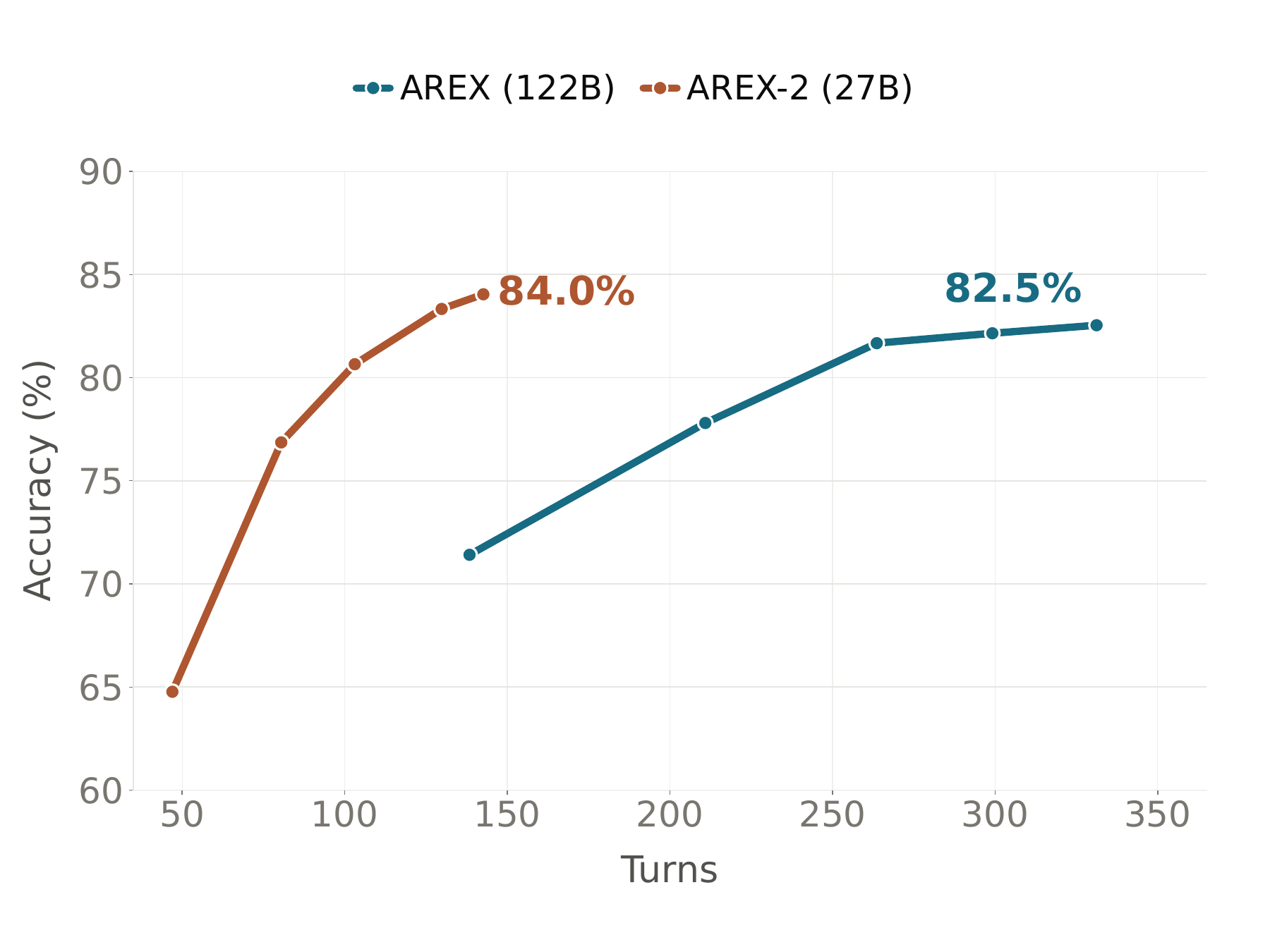}
    \captionof{figure}{
      Test-time scaling for capability transfer on BrowseComp, showing
  accuracy versus average turns with no correctness feedback provided
  during evaluation.
    }
    \label{fig:browsecomp-scaling}
  \end{minipage}

\end{figure*}

\paragraph{Setup.} We use the 188 problems of the Frontier-CS Agent Track, with a budget of five hours per problem, measured from the start of the agent session. The agent works on each problem in rounds. In a round, it implements or revises a solution, compiles it, runs the provided sample cases, and writes additional randomized and edge-case tests of its own. It then submits the solution and receives a score with evaluation feedback, which it uses to decide what to change in the next round. AREX-2 runs with our optimized harness. \Cref{fig:fcs-main} reports the mean best-so-far score over time.

\paragraph{Results.} AREX-2 improves throughout the five hours. It does not start with the highest score, but it takes the lead as the budget grows, reaching 54.4 after one hour, 65.9 after two hours, and 70.7 after five hours. In contrast, DeepSeek-V4-Pro stops improving at 44.7 after two hours, and DeepSeek-V4-Flash at 39.1 after three hours. For AREX-2 the later hours remain productive: it gains 4.8 points between the second and the fifth hour, of which 2.2 points come in the final hour.

\paragraph{What it verifies.} The advantage of AREX-2 comes from sustained improvement, and not only from a strong start. Its gains become smaller over time but remain positive until the end of the budget, which corresponds to a long effective horizon $T^{\ast}$. The two baselines reach their effective horizon within two to three hours, after which a larger budget no longer helps them.

\subsubsection{Round Scaling without Feedback}
\label{sec:case-browsecomp}

\paragraph{Setup.} We use BrowseComp, a deep-research benchmark for which no new trajectories were added in training. Unlike in the previous setting, the agent receives no score during a run, so it never learns whether its current answer is correct. AREX-2 follows a recursive research process with two loops. The inner loop gathers evidence and produces a provisional answer. The outer loop decides whether to accept the answer, refine it, or restart, based on the model's own confidence and the state of the trajectory. We vary the budget, measured as the average number of turns per question, where a turn is one LLM call, and report accuracy in \Cref{fig:browsecomp-scaling}. For comparison, we run \textsc{AREX} (122B), trained with the previous recipe, under the same process.

\paragraph{Results.} The accuracy of AREX-2 rises steadily with the budget, from 64.8 at 47 turns to 84.0 at 143 turns, with diminishing returns as the budget grows. \textsc{AREX} (122B) also improves with more turns, but far less efficiently. At a similar budget of about 140 turns, AREX-2 is more than 12 points ahead. It also exceeds the final accuracy of \textsc{AREX} with less than half of the turns, and its gain per turn is about three times as large. AREX-2 achieves this with 27B parameters, less than a quarter of the size of \textsc{AREX}.

\paragraph{What it verifies.} AREX-2 can improve its answer without an external correctness signal. The gains come from the model continuing to search, reassess, and refine by its own judgment, and not from correcting an answer after being told that it was wrong. Its gain per turn is also about three times that of \textsc{AREX}, which corresponds to a larger $\bar r$ in \Cref{sec:formalization}. Since the two models share the same deep-research training data and the new trajectories contain no research tasks, this difference is evidence that long-horizon reflection transfers across domains.

\subsubsection{Stage-wise Ablation}
\label{sec:case-mle}

\begin{figure}[t]
  \centering
  \includegraphics[width=0.95\linewidth]{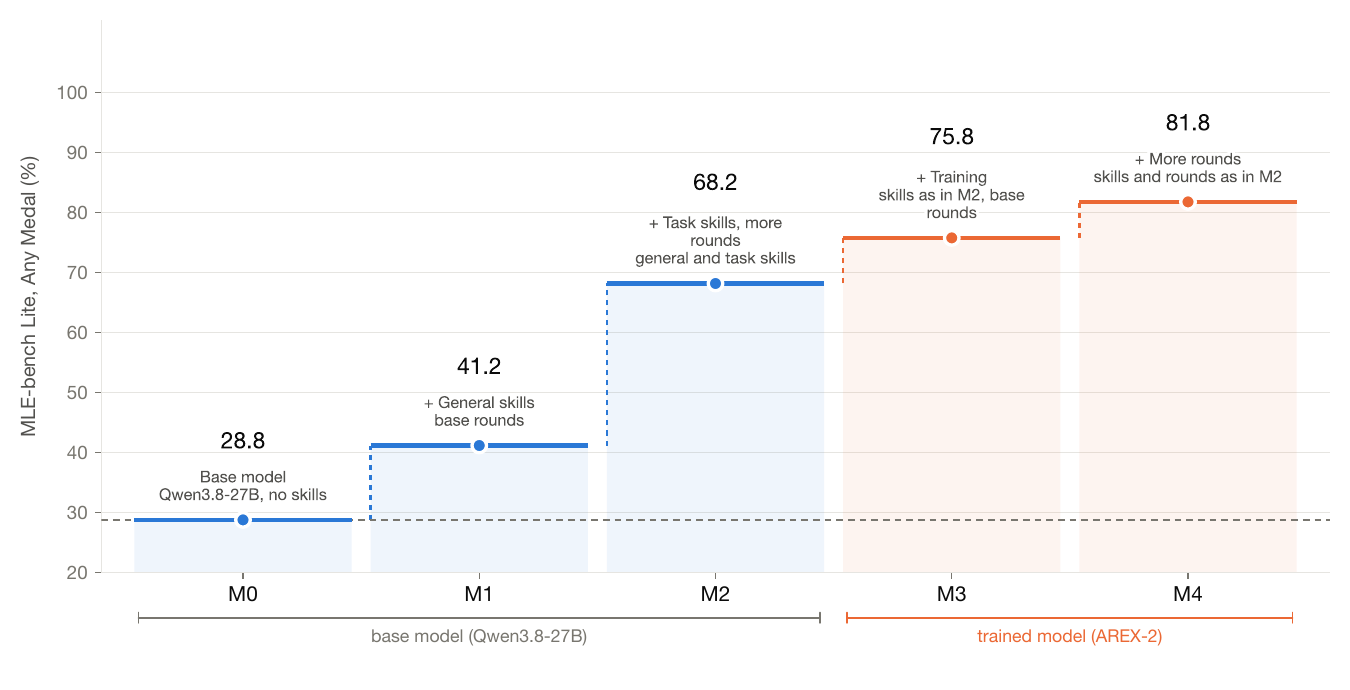}
  \caption{Stage-wise ablation on MLE-bench Lite. Stages M0 to M2 use the base model, and M3 and M4 use AREX-2.}
  \label{fig:mle-ladder}
\end{figure}

\paragraph{Setup.} We measure how much each part of our approach contributes, in five stages that are all evaluated on MLE-bench Lite (\Cref{fig:mle-ladder}). The stages vary three factors: the model, the skills in its context, and the number of rounds. Stages M0 to M2 use the base model, Qwen3.8-27B. M0 has no skills, M1 adds general skills, and M2 adds task-specific skills and more rounds. Stages M3 and M4 use the trained model, AREX-2, with the same skills as M2. M3 uses the same number of rounds as M0 and M1, and M4 uses the same number as M2. General skills are prepared in advance from machine learning repositories and common practices. Task-specific skills are built by searching for information related to a task, and material about the task itself is excluded.

\paragraph{Results.} The base model scores 28.8. General skills raise the score to 41.2, and task-specific skills with more rounds raise it to 68.2. With the same skills, AREX-2 reaches 75.8 with the base number of rounds and 81.8 with more rounds.

\paragraph{What it verifies.} The ablation supports three observations. First, operational knowledge matters: skills and more rounds raise the base model from 28.8 to 68.2 without any training. Second, training adds to what skills provide. M2 and M4 have the same skills and the same number of rounds and differ only in the model, and AREX-2 exceeds the base model by 13.6 points. AREX-2 is ahead even with fewer rounds: M3 exceeds M2 by 7.6 points. Third, the trained model turns more rounds into further improvement, gaining 6.0 points from M3 to M4, which is consistent with the scaling results above.

\section{Conclusion}
\label{sec:conclusion}

We set out to move the improvement loop inside the model. We
defined self-improvement as turning more rounds on a task into a better
solution, and traced it to two complementary capabilities: reflection
raises the gain per round, and long-horizon execution keeps the rounds
productive. Neither is learned from finished solutions, so we built data
that records the work: trajectories in which an agent improves a
solution over many rounds, in environments compiled from machine
learning repositories and algorithmic problems.

Trained on them, AREX-2 achieves leading results on MLE-bench Lite and
Frontier-CS, and improves on deep research as well, for which no new
data was added. It keeps improving as its budget grows: on Frontier-CS
through the fifth hour, and on BrowseComp without any correctness
signal. A stage-wise ablation further shows that operational
knowledge, training, and a larger budget each contribute to the result. Long-horizon reflection, learned where it can be
supervised, carries over beyond the domains it was learned in. The path
forward is to widen the training domains, lengthen the horizons, and
close the loop by letting the model's own trajectories become its next
training data.
\definecolor{contributionblue}{RGB}{0,0,120}

\section*{Contributions}

\begingroup
\setlength{\parindent}{0pt}
\setlength{\parskip}{0pt}
\raggedright

\textbf{Core Contributors:}
Hongjin Qian\footnote{These authors contributed equally to this work.}, \
Chaofan Li\footnotemark[1], \
Kun Luo\footnotemark[1], \
Wenqing Wei\footnotemark[1], \
Jianlyu Chen\footnotemark[1], \
Zheng Liu\footnote{Zheng Liu is the project leader.}

\vspace{0.5em}

\textbf{Participants:}
Shuqi Lu, \
Yuyang Hu, \
Hongwang Xiao, \
Hui Wang, \
Chaozhuo Li, \
Qiwei Ye

\vspace{0.5em}

\textbf{Advisors:}
Zhicheng Dou, \
Defu Lian

\vspace{1.5em}

\bibliographystyle{assets/plainnat}
\bibliography{citation}

\end{document}